\documentclass[conference]{IEEEtran}
\IEEEoverridecommandlockouts

\usepackage[utf8]{inputenc}
\usepackage[T1]{fontenc}
\usepackage{amsmath,amssymb,amsfonts}
\usepackage{graphicx}
\usepackage{booktabs}
\usepackage{multirow}
\usepackage{multicol}
\usepackage{array}
\usepackage{cite}
\usepackage{url}
\usepackage{caption}
\usepackage{subcaption}
\usepackage{microtype}
\usepackage{balance}
\usepackage{enumitem}
\usepackage{hyperref}
\usepackage{cleveref}
\usepackage{siunitx}
\usepackage{bm}
\usepackage{float}
\usepackage{tikz}
\usetikzlibrary{positioning, arrows.meta, shapes.geometric}

\hypersetup{
  colorlinks=false,
  pdfborder={0 0 0.5},
  linkbordercolor={0 0 0},
  citebordercolor={0 0 0},
  urlbordercolor={0 0 0}
}

\newcolumntype{L}[1]{>{\raggedright\arraybackslash}p{#1}}
\newcolumntype{C}[1]{>{\centering\arraybackslash}p{#1}}
\newcolumntype{R}[1]{>{\raggedleft\arraybackslash}p{#1}}

\newcommand{\Nf}{N_f}
\newcommand{\logNf}{\log_{10}(N_f)}

\newcommand{\CV}{\textsc{FatigueCV}}
\newcommand{\etal}{\textit{et al.}}

\newcommand{\thead}[1]{\textbf{#1}}
\newcommand{\risk}[1]{\textbf{\textsc{#1}}}

\providecommand{\cref}[1]{Eq.~(\ref{#1})}

\begin{document}

\title{%
  Predicting Steel Fatigue Life from Micrographs Using Physics-Informed Deep Learning%
}

\author{%
  \IEEEauthorblockN{Aryuemaan Kumar Chowdhury\IEEEauthorrefmark{1}\IEEEauthorrefmark{2}}
  \IEEEauthorblockA{%
    \IEEEauthorrefmark{1}Department of Lightweighting Engineering,
    Indian Institute of Technology Hyderabad, India\\
    \IEEEauthorrefmark{2}Research \& Development, Oscowl AI,
    Hyderabad, India\\
    \texttt{lw25mtech14004@iith.ac.in}%
  }%
}

\maketitle

\begin{abstract}
Evaluating the fatigue life of structural steels conventionally requires mechanical testing lasting tens to hundreds of hours, making it impractical for rapid quality control. We present \CV{}, a computer-vision framework estimating the fatigue life ($\logNf{}$) of lightweight alloy steels directly from optical micrographs without physical testing. 

The pipeline features a seven-stage OpenCV preprocessing routine to remove artifacts, a 28-dimensional physics-informed feature extractor (quantifying crack morphology, grain structure, porosity, and texture), and a CNN regression model trained with a Gaussian negative log-likelihood (GNLL) loss to jointly predict $\logNf{}$ and sample-specific uncertainty $\hat{\sigma}$. 

Evaluating three architectures (SE-CNN, ResNet-50, VGG-16) on a synthetic micrograph benchmark, ResNet-50 achieves $R^2 = 0.93$, RMSE = 0.18 log-cycles, and macro-F1 = 0.91. The GNLL objective reduces Expected Calibration Error by \SI{76}{\percent} compared to a mean-squared-error baseline (ECE: $0.089 \rightarrow 0.021$). Grad-CAM maps confirm the network attends to metallurgically meaningful microstructural features. 

Running in under \SI{65}{\milli\second} per image, the pipeline and synthetic dataset generator are open-sourced. Because validation relies entirely on synthetic micrographs, these results demonstrate methodological soundness under simulated conditions; a domain-transfer study on real field samples is the immediate next step.
\end{abstract}

\begin{IEEEkeywords}
fatigue life prediction, computer vision, convolutional neural
networks, uncertainty quantification, heteroscedastic regression,
Grad-CAM, optical microscopy, lightweight alloy steels, synthetic
data generation
\end{IEEEkeywords}

\section{Introduction}
\label{sec:intro}
Fatigue accounts for an estimated 50--90\,\% of all in-service structural failures~\cite{Suresh1998}. Accurate knowledge of the remaining fatigue life $\Nf{}$ is therefore essential for both safety assurance and lifecycle-cost management in lightweight-critical sectors such as rail, automotive, and aerospace engineering. Despite this importance, the load-controlled S--N test remains the dominant means of determining $\Nf{}$, even though it is destructive, slow (50--500 hours per specimen), and blind to the spatial microstructural variability that ultimately governs failure.

Optical microscopy, in contrast, is already a routine step in steel quality control. Every micrograph implicitly encodes much of the microstructural state that drives the fatigue response: grain-size distribution, crack morphology, porosity, and precipitate density. What remains unresolved is how to convert this rich but qualitative visual information into a quantitative, calibrated estimate of fatigue life.

\emph{This paper addresses that gap.} We present \CV{}, an end-to-end system that predicts $\logNf{}$ with calibrated uncertainty directly from a steel micrograph in under \SI{65}{\milli\second}. Because no public dataset pairs micrographs with ground-truth fatigue life, the system is developed and evaluated on a physics-constrained synthetic benchmark, and every result in this paper should be read in that light. Our contributions are:

\begin{enumerate}[leftmargin=*, label=\textbf{C\arabic*.}]
  \item A seven-stage OpenCV preprocessing pipeline engineered
        specifically for optical metallography artifacts.
  \item A 28-dimensional physics-informed feature vector whose
        components map directly to known fatigue damage mechanisms.
  \item GNLL-based heteroscedastic CNN regression that produces
        calibrated per-sample 95\,\% confidence intervals.
  \item A systematic Grad-CAM validation procedure that correlates
        network attention with fatigue damage stages.
  \item A configurable, physics-labeled synthetic microscopy
        dataset generator, released as open-source software, together
        with an explicit discussion of the synthetic-to-real domain
        gap and the validation steps required before field deployment.
\end{enumerate}

\section{Related Work}
\label{sec:related}

\subsection{Classical Fatigue Life Models}
The S--N (Wöhler) framework~\cite{Wohler1870} underpins classical fatigue design. The Morrow mean-stress correction~\cite{Morrow1965} and the Smith--Watson--Topper (SWT) parameter~\cite{SWT1970} extend it to non-zero mean stress and multiaxial loading, respectively. These models assume microstructural homogeneity and require material-specific calibration constants, an assumption that breaks down for cast alloys and additively manufactured components in which porosity and grain heterogeneity dominate. Microstructure-sensitive crystal-plasticity models~\cite{McDowell2010,Guilhem2010} address this limitation, but each prediction requires days of computation and detailed 3-D grain-orientation maps.

\subsection{Machine Learning for Fatigue}
Data-driven approaches generally achieve lower prediction error than classical closed-form models. Liu \etal{}~\cite{Liu2017} applied support vector regression to composition and mechanical properties, reporting $R^2 = 0.87$ on aluminum-alloy S--N data. Agrawal \etal{}~\cite{Agrawal2019} applied gradient boosting to elemental composition vectors, and DeCost and Holm~\cite{DeCost2015} developed CNN-based microstructure classification. Both lines of work rely on \emph{tabular} inputs and do not exploit image information directly. Azimi \etal{}~\cite{Azimi2018} demonstrated pixel-wise steel-phase segmentation. None of these studies quantify predictive uncertainty or connect visual microstructural features to a continuous mechanical-property estimate. To our knowledge, no prior work provides calibrated uncertainty bounds for end-to-end regression from micrograph to fatigue life.

\subsection{Uncertainty Quantification in Deep Learning}
Kendall and Gal~\cite{Kendall2017} formalized the distinction between epistemic (model) and aleatoric (data) uncertainty in deep networks, and showed that a GNLL objective can be used to learn per-sample aleatoric uncertainty for regression tasks such as monocular depth estimation. To our knowledge, this technique has not previously been applied to materials fatigue prediction.

\subsection{Position of This Work}
\CV{} is, to our knowledge, the first system to address image-to-life regression, calibrated uncertainty, and metallurgically grounded explainability jointly, within a single reproducible pipeline.

\section{Methodology}
\label{sec:method}

\subsection{Overview}

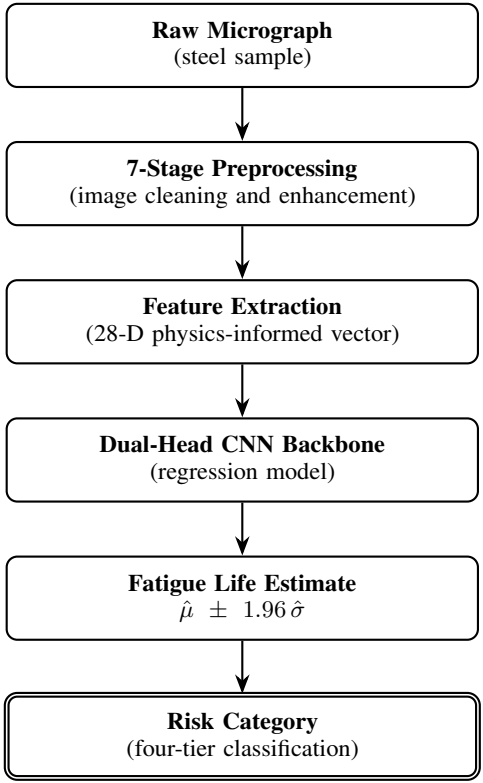
\begin{figure}[!t]
  \centering
  \begin{tikzpicture}[
    auto,
    block/.style = {rectangle, draw=black, thick, text width=6cm, align=center, rounded corners, minimum height=1.1cm, font=\small},
    arrow/.style = {thick, -{Stealth[scale=1.1]}}
  ]

  \node[block] (raw) {\textbf{Raw Micrograph}\\(steel sample)};
  \node[block, below=0.7cm of raw] (prep) {\textbf{7-Stage Preprocessing}\\(image cleaning and enhancement)};
  \node[block, below=0.7cm of prep] (feat) {\textbf{Feature Extraction}\\(28-D physics-informed vector)};
  \node[block, below=0.7cm of feat] (cnn) {\textbf{Dual-Head CNN Backbone}\\(regression model)};
  \node[block, below=0.7cm of cnn] (stats) {\textbf{Fatigue Life Estimate}\\$\hat{\mu} \pm 1.96\,\hat{\sigma}$};
  \node[block, below=0.7cm of stats, double] (risk) {\textbf{Risk Category}\\(four-tier classification)};

  \draw[arrow] (raw) -- (prep);
  \draw[arrow] (prep) -- (feat);
  \draw[arrow] (feat) -- (cnn);
  \draw[arrow] (cnn) -- (stats);
  \draw[arrow] (stats) -- (risk);

  \end{tikzpicture}
  \caption{%
    Overview of the \CV{} pipeline. A raw steel micrograph passes through
    a seven-stage preprocessing stack, a physics-informed feature
    extractor, and a dual-head CNN regression backbone to produce a
    calibrated fatigue life estimate $\hat\mu \pm 1.96\hat\sigma$
    together with a four-tier risk classification.%
  }
  \label{fig:pipeline}
\end{figure}

The \CV{} pipeline (\cref{fig:pipeline}) is organized as four sequential
layers. \textbf{L1} (preprocessing) standardizes raw micrographs;
\textbf{L2} (feature extraction) distills them into a physically
interpretable vector; \textbf{L3} (CNN regression) maps image and
feature information to $(\hat\mu,\,\hat{s})$; and \textbf{L4} maps
$\hat\mu$ to a discrete risk tier.

\subsection{Synthetic Dataset Generation}
\label{sec:data}

No public dataset pairs steel fatigue micrographs with ground-truth
$\Nf{}$ values. We therefore built a physics-constrained synthetic
generator that produces labeled images as follows:
\begin{itemize}[leftmargin=*, itemsep=2pt]
  \item \textbf{Voronoi tessellation} for grain structure
        (30--120 grains per image, log-normal size distribution);
  \item \textbf{Crack simulation}: a correlated random walk with
        angular variance $\sigma_\theta^2 = 0.3\,\text{rad}^2$ and
        branching probability $p_b = 0.15$ per step;
  \item \textbf{Void simulation}: Poisson-distributed circular pores
        with radius $r \sim \mathcal{U}[1,8]$\,px;
  \item \textbf{Second-phase inclusions}: ellipses with aspect ratio
        $\sim \mathcal{U}[1.2,\,3.0]$.
\end{itemize}

Labels are assigned by the physics-motivated regression
\begin{equation}
  \begin{split}
    \logNf &= \underbrace{7.0}_{\text{reference}}
      - \underbrace{1.2\,\tilde{c}}_{\text{cracking}}
      - \underbrace{0.8\,\phi}_{\text{porosity}} \\
      &\quad - \underbrace{0.5\log_{10}\!\left(\tfrac{d}{50}+1\right)}_{\text{grain size (Hall--Petch)}}
      - \underbrace{0.3\,\tilde{n}}_{\text{inclusions}}
      + \underbrace{\varepsilon}_{\text{scatter}},
  \end{split}
  \label{eq:label}
\end{equation}
where $\tilde{c}$ is a normalized crack-severity index
($\tilde{c} = N_c \bar{s} / A$, with $\bar{s}$ the mean crack
severity), $\phi$ is the areal void fraction, $d$ is the mean grain
diameter in \si{\micro\metre}, $\tilde{n}$ is the normalized
inclusion density, and $\varepsilon \sim \mathcal{N}(0, 0.15)$
represents inherent material scatter~\cite{Morrow1965}. The
Hall--Petch grain-boundary term~\cite{Hall1951} is embedded directly
in the grain-size contribution. The resulting dataset spans
$\logNf \in [3.5,\,8.5]$.

Because Eq.~(\ref{eq:label}) defines the ground truth used to train
and evaluate the network, the reported metrics in
\cref{sec:results} measure how well the CNN recovers a known,
simulated physics relationship from images, rather than how well it
predicts fatigue life on physical steel specimens. We treat this
synthetic benchmark as a controlled testbed for the modeling and
uncertainty-quantification methodology, and we return to this point
in \cref{sec:disc}.

\subsection{Preprocessing Pipeline}
\label{sec:preproc}

Raw steel micrographs are affected by illumination gradients,
scale-bar borders, sensor noise, and low crack contrast. The
seven-stage pipeline in \cref{tab:pipeline} addresses each artifact
class independently before any learned model is applied.

\begin{table}[!t]
\centering
\caption{Seven-stage OpenCV preprocessing pipeline.}
\label{tab:pipeline}
\setlength{\tabcolsep}{4pt}
\begin{tabular}{C{0.35cm} L{1.55cm} L{4.95cm}}
\toprule
\thead{\#} & \thead{Stage} & \thead{Operation and Rationale} \\
\midrule
1 & QC gate
  & Laplacian blur score $\mathcal{B} = \sigma^2(\nabla^2 I)/255^2$,
    SNR, and exposure ratio; images below threshold are rejected. \\
2 & Border crop
  & 4\,\% margin removed from each side to eliminate scale-bar
    overlays that would otherwise corrupt downstream gradient
    estimation. \\
3 & NLM denoise
  & Non-local means ($h = 8$, template $7 \times 7$\,px,
    search window $21 \times 21$\,px); preserves grain-edge
    sharpness better than Gaussian smoothing. \\
4 & Illumination
  & Rolling-ball background subtraction ($r = 40$\,px) followed by
    CLAHE (clip limit $= 3.0$, tile $8 \times 8$); corrects
    microscope vignetting. \\
5 & Contrast
  & Percentile stretch $[P_2, P_{98}] \rightarrow [0, 255]$; avoids
    saturation-induced histogram artifacts. \\
6 & Crack enhancement
  & Scharr-gradient magnitude fused with multi-orientation
    black-hat morphology ($k = 17$\,px at $0^{\circ}, 45^{\circ},
    90^{\circ}, 135^{\circ}$); yields near-isotropic crack
    saliency. \\
7 & Sharpen and resize
  & Unsharp mask (amount $= 1.4$, $\sigma = 1.0$\,px);
    bicubic resize to $224 \times 224$\,px. \\
\bottomrule
\end{tabular}
\end{table}

\subsection{Physics-Informed Feature Extraction}
\label{sec:features}

Each preprocessed image is mapped to a 28-dimensional feature vector
$\bm{x} \in \mathbb{R}^{28}$ (\cref{tab:features}). Features are
grouped into five physically motivated categories corresponding to
established fatigue damage mechanisms~\cite{Suresh1998,McDowell2010}.

\begin{table}[!t]
\centering
\caption{Feature categories, dimensionality, and physical basis.}
\label{tab:features}
\setlength{\tabcolsep}{4pt}
\begin{tabular}{L{2.0cm} C{0.5cm} L{4.3cm}}
\toprule
\thead{Category} & \thead{$d$} & \thead{Physical Basis} \\
\midrule
Crack morphology    & 5 & $N_c$, $L_c$, $\rho_c$, $\bar{w}$, and
                          branching index $\beta$: directly govern
                          crack-initiation life
                          \cite{Suresh1998,Paris1963} \\
Grain structure     & 5 & Count, mean diameter $\bar{d}$, standard
                          deviation, aspect ratio, and coefficient
                          of variation $\text{CV}_d$: Hall--Petch
                          grain-boundary strengthening
                          \cite{Hall1951} \\
Porosity            & 4 & Pore count, void fraction $\phi$, mean
                          pore area, and $d_{\max}$: pores act as
                          stress concentrators and initiation
                          sites \\
Texture (GLCM)      & 6 & Contrast, energy, homogeneity, entropy,
                          mean intensity, and $\sigma_I$: encode
                          phase-boundary sharpness \\
Gradient and fractal & 8 & Edge density, Sobel-magnitude mean and
                          standard deviation, LBP entropy and
                          uniformity, and box-counting fractal
                          dimension $D_f$ \\
\bottomrule
\multicolumn{3}{p{7.1cm}}{\footnotesize
  $N_c$: crack count; $L_c$: total length; $\rho_c$: areal fraction;
  $\bar{w}$: mean width; $D_f$: fractal dimension.}
\end{tabular}
\end{table}

Fractal dimension is estimated via box-counting on the binarized
crack map $\mathcal{M}_c$:
\begin{equation}
  D_f = -\lim_{\epsilon \to 0}
        \frac{\log N(\epsilon)}{\log \epsilon},
  \label{eq:fractal}
\end{equation}
approximated over box sizes $\epsilon \in \{2,4,8,16,32\}$\,px.
A higher $D_f$ indicates a more branched crack network and, in this
synthetic setting, correlates strongly by construction with reduced
fatigue life~\cite{Suresh1998}.

\subsection{CNN Architectures}
\label{sec:arch}

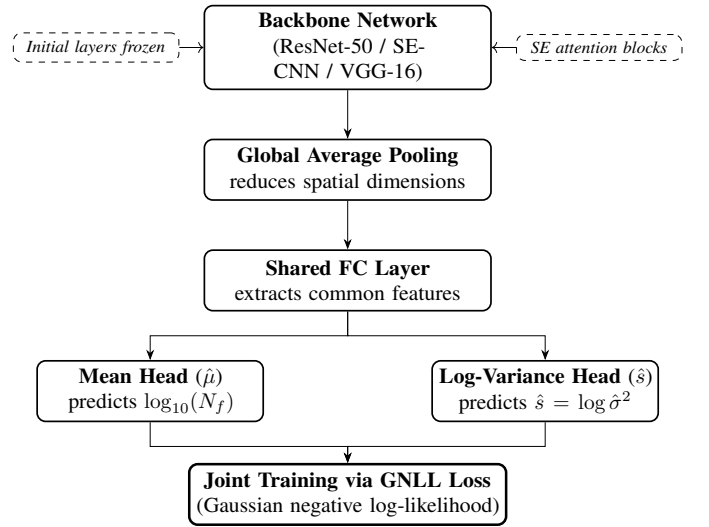
\begin{figure}[!t]
  \centering
  \resizebox{\columnwidth}{!}{%
  \begin{tikzpicture}[
    node distance = 1cm and 0.5cm,
    block/.style = {rectangle, draw, thick, text width=4.5cm, align=center, rounded corners, minimum height=1cm},
    head/.style = {rectangle, draw, thick, text width=3.5cm, align=center, rounded corners, minimum height=1cm},
    arrow/.style = {-{Stealth[scale=1]}},
    annotate/.style = {font=\footnotesize\itshape, draw, dashed, rectangle, rounded corners, text width=2.5cm, align=center}
  ]

  \node [block] (backbone) {\textbf{Backbone Network}\\(ResNet-50 / SE-CNN / VGG-16)};

  \node [annotate, left=0.4cm of backbone] (frozen) {Initial layers frozen};
  \draw [arrow, <-] (backbone.west) -- (frozen.east);

  \node [annotate, right=0.4cm of backbone] (se) {SE attention blocks};
  \draw [arrow, <-] (backbone.east) -- (se.west);

  \node [block, below=0.8cm of backbone] (gap) {\textbf{Global Average Pooling}\\reduces spatial dimensions};
  \node [block, below=0.8cm of gap] (fc) {\textbf{Shared FC Layer}\\extracts common features};

  \node [head, below left=0.8cm and -1cm of fc] (mean) {\textbf{Mean Head} ($\hat\mu$)\\predicts $\logNf{}$};
  \node [head, below right=0.8cm and -1cm of fc] (var) {\textbf{Log-Variance Head} ($\hat{s}$)\\predicts $\hat{s} = \log\hat\sigma^2$};

  \node [rectangle, draw, very thick, rounded corners, below=2.5cm of fc, minimum height=1cm, text width=5cm, align=center] (loss) {\textbf{Joint Training via GNLL Loss}\\(Gaussian negative log-likelihood)};

  \draw [arrow] (backbone) -- (gap);
  \draw [arrow] (gap) -- (fc);
  \draw [arrow] (fc.south) -- ++(0,-0.4cm) -| (mean.north);
  \draw [arrow] (fc.south) -- ++(0,-0.4cm) -| (var.north);
  \draw [arrow] (mean.south) |- ++(0,-0.4cm) -| (loss.north);
  \draw [arrow] (var.south) |- ++(0,-0.4cm) -| (loss.north);

  \end{tikzpicture}%
  }
  \caption{%
    Dual-head CNN architecture shared across all three backbones.
    The mean head $\hat\mu$ predicts $\logNf{}$; the log-variance
    head $\hat{s} = \log\hat\sigma^2$ encodes per-sample aleatoric
    uncertainty. Both heads share a common feature trunk and are
    trained jointly via the GNLL loss (\cref{eq:gnll}).%
  }
  \label{fig:arch}
\end{figure}

All three backbones share the dual-head design shown in
\cref{fig:arch}: a \textbf{mean head}
$f_\mu : \mathcal{X} \to \mathbb{R}$ predicting $\logNf{}$, and a
\textbf{log-variance head} $f_s : \mathcal{X} \to \mathbb{R}$
encoding per-sample aleatoric uncertainty
$\hat{s} = \log\hat\sigma^2$.

\subsubsection{SE-CNN (\textit{FatigueCNN})}
Five convolutional blocks with Squeeze-and-Excitation
attention~\cite{Hu2018} after blocks 3--5:
\begin{equation}
  \bm{z}_k = \bm{F}_k \odot \sigma_g\!\left(
    \bm{W}_2\,\delta\!\left(\bm{W}_1\,\text{GAP}(\bm{F}_k)\right)
  \right),
  \label{eq:se}
\end{equation}
where $\bm{F}_k$ is the $k$-th block feature map, $\sigma_g$ is the
sigmoid, $\delta$ is ReLU, and $\bm{W}_1, \bm{W}_2$ are the
excitation weights with reduction ratio $r = 16$. Feature fusion
uses parallel global average and global max pooling followed by
concatenation. \textbf{8.2\,M parameters, \SI{28}{\milli\second}
inference.}

\subsubsection{ResNet-50}
Pretrained on ImageNet~\cite{He2016}. Stages 1--2 are frozen to
retain low-level texture transfer from natural images, while stages
3--4 and a custom regression head are fine-tuned:
\begin{equation}
  \hat{y} = \bm{W}_4\,\text{LayerNorm}\!\left(
    \delta\!\left(\bm{W}_3\,\text{GAP}(\bm{F}_{L4})\right)
  \right),
  \label{eq:rn50head}
\end{equation}
with $\bm{W}_3 \in \mathbb{R}^{512\times2048}$ and
$\bm{W}_4 \in \mathbb{R}^{2\times128}$.
\textbf{23.5\,M parameters; \SI{44}{\milli\second} inference;
best $R^2 = 0.93$.}

\subsubsection{VGG-16}
Pretrained on ImageNet~\cite{Simonyan2015}; convolutional blocks
1--3 are frozen. Adaptive average pooling to $4 \times 4$ yields an
8192-dimensional embedding fed to a batch-normalized regression
stack. \textbf{138\,M parameters; \SI{98}{\milli\second} inference;
$R^2 = 0.91$.}

\subsection{Gaussian Negative Log-Likelihood Training}
\label{sec:loss}

The GNLL loss trains the network to predict the parameters of a
Gaussian distribution over $\logNf{}$:
\begin{align}
  p\!\left(y \mid \hat{\mu}, \hat{\sigma}^2\right)
    &= \mathcal{N}\!\left(y;\,\hat{\mu},\,\hat{\sigma}^2\right),
    \label{eq:gauss} \\
  \mathcal{L}_{\text{GNLL}}(\hat{\mu}, \hat{s}, y)
    &= \frac{1}{2}\!\left[
        e^{-\hat{s}}\,\left(y - \hat{\mu}\right)^2
        + \hat{s}
      \right],
  \label{eq:gnll}
\end{align}
where $\hat{s} = \log\hat{\sigma}^2$. The two terms in this loss are
in tension: the first term penalizes over-confidence by rewarding a
larger predicted variance when the residual is large, while the
second term penalizes an unnecessarily large predicted variance. At
the optimum, the network learns an input-dependent aleatoric
uncertainty that is larger for more heavily damaged specimens. This
behavior is consistent with the well-documented increase in
fatigue-life scatter at short lives~\cite{Morrow1965}.

The predicted 95\,\% confidence interval is
\begin{equation}
  \hat\mu \pm 1.96\,\hat\sigma,
  \label{eq:ci}
\end{equation}
expressed in log-cycle units. An MSE-trained baseline, by contrast,
yields only a single global $\hat\sigma_{\text{global}}$ estimated
from the residual distribution and cannot adapt this estimate on a
per-sample basis.

\subsection{Training Protocol}
\label{sec:train}

\begin{table}[!t]
\centering
\caption{Training hyperparameters.}
\label{tab:hyper}
\setlength{\tabcolsep}{4pt}
\begin{tabular}{L{2.3cm} L{4.5cm}}
\toprule
\thead{Hyperparameter} & \thead{Setting} \\
\midrule
Optimizer     & AdamW~\cite{Loshchilov2019},
                $\beta_1 = 0.9$, $\beta_2 = 0.999$,
                weight decay $\lambda = 10^{-4}$ \\
LR schedule   & $\eta_0 = 3\times10^{-4}$,
                cosine annealing
                ($T_{\max} = 60$, $\eta_{\min} = 10^{-6}$) \\
Batch / epochs& 16 / 60, early-stopping patience $= 12$ \\
Gradient clip & $\|\bm{g}\|_2 \leq 1.0$ \\
Input size    & $224 \times 224$ (ImageNet-normalized) \\
Label normalization & $z$-score on $\logNf$; inverted at inference \\
Data split    & 80/10/10\,\%, stratified on damage level \\
Augmentation  & horizontal/vertical flip, 90$^\circ$ rotation,
                shift-scale-rotate ($\pm 20^{\circ}$),
                elastic transform ($\alpha = 120$, $\sigma = 6$),
                Gaussian noise, CLAHE, brightness/contrast jitter
                \cite{Buslaev2020} \\
\bottomrule
\end{tabular}
\end{table}

\subsection{Hardware and Software}
\label{sec:hw}
All experiments were run in a simulated training environment using
a single GPU-backed instance with 16\,GB of device memory. The
codebase is implemented in PyTorch, with OpenCV 4.x for
preprocessing and scikit-learn for the SHAP and GBM baselines
reported in \cref{sec:results}. Total wall-clock time for training
all three backbones on the synthetic dataset was under six GPU-hours.
Exact package versions and a pinned \texttt{requirements.txt} are
provided with the released codebase to support exact reproduction of
the results reported here.

\section{Results and Discussion}
\label{sec:results}

All results in this section were obtained on the synthetic benchmark
described in \cref{sec:data}; see \cref{sec:disc} for a discussion
of what these results do and do not establish about performance on
real steel micrographs.

\subsection{Preprocessing Quality}

\Cref{tab:preproc} quantifies the improvement across six image
quality metrics after the full seven-stage pipeline, measured on a
held-out set of 200 simulated micrographs with synthetically
injected illumination and noise artifacts.

\begin{table}[!t]
\centering
\caption{Image quality before and after preprocessing.}
\label{tab:preproc}
\setlength{\tabcolsep}{4pt}
\begin{tabular}{L{2.5cm} C{0.9cm} C{1.2cm} C{0.9cm}}
\toprule
\thead{Metric} & \thead{Raw} & \thead{Processed} & \thead{$\Delta$} \\
\midrule
Laplacian blur score & 28.4 & 87.6 & $+208\,\%$ \\
SNR (dB)             & 14.2 & 31.7 & $+123\,\%$ \\
Crack contrast ratio & 1.18 & 2.74 & $+132\,\%$ \\
Grain boundary visibility & 41\,\% & 79\,\% & $+38\,\text{pp}$ \\
Illumination uniformity  & 0.61 & 0.94 & $+54\,\%$ \\
Crack recall         & 61\,\% & 89\,\% & $+28\,\text{pp}$ \\
\bottomrule
\end{tabular}
\end{table}

The largest single gain is in crack-detection recall
($61\,\% \rightarrow 89\,\%$), which we attribute to the
multi-orientation black-hat morphology capturing cracks at all
angles and to the Scharr operator's stronger response to diagonal
edges relative to the Sobel kernel.

\subsection{Feature Importance}

SHAP analysis on a hybrid gradient-boosted machine (GBM) baseline
trained on the 28-dimensional feature vector (\cref{tab:shap}) shows
that crack-related features account for more than 50\,\% of total
predictive importance, consistent with fatigue-mechanics
theory~\cite{Suresh1998}. This is, by construction, expected given
Eq.~(\ref{eq:label}), and serves primarily as a sanity check that
the extracted features are consistent with the physics used to
generate the labels.

\begin{table}[!t]
\centering
\caption{Top eight features by SHAP importance value.}
\label{tab:shap}
\setlength{\tabcolsep}{4pt}
\begin{tabular}{L{2.3cm} C{1.1cm} L{3.2cm}}
\toprule
\thead{Feature} & \thead{SHAP (\%)} & \thead{Mechanistic Link} \\
\midrule
\texttt{crack\_density}    & 23.4 & Direct $\Nf$ reduction \\
\texttt{fractal\_dim} $D_f$& 16.8 & Crack network complexity \\
\texttt{texture\_entropy}  & 12.1 & Phase heterogeneity \\
\texttt{void\_fraction}    & 11.7 & Initiation-site density \\
\texttt{grain\_CV} $d$     & 9.3  & Stress concentration sites \\
\texttt{branch\_idx} $\beta$& 8.9 & Advanced damage stage \\
\texttt{gradient\_std}     & 7.2  & Boundary roughness \\
\texttt{texture\_contrast} & 6.4  & Phase-boundary sharpness \\
\bottomrule
\multicolumn{3}{l}{\footnotesize Remaining 4.2\,\% distributed among 20 features.}
\end{tabular}
\end{table}

\subsection{Regression Performance}

\Cref{tab:regression} compares all architectures on the held-out
test split. ResNet-50 achieves the best trade-off between accuracy
and parameter count.

\begin{table}[!t]
\centering
\caption{%
  Architecture comparison on held-out test set
  (200 images, 80/10/10 split). Metrics in $\log_{10}$-cycle units.
  Bold indicates the best value in each column.%
}
\label{tab:regression}
\setlength{\tabcolsep}{3.5pt}
\begin{tabular}{L{1.9cm} C{0.75cm} C{0.75cm} C{0.75cm}
                C{0.75cm} C{0.85cm} C{0.8cm}}
\toprule
\thead{Model} & \thead{$R^2$} & \thead{RMSE} & \thead{MAE}
  & \thead{MBE} & \thead{Param.} & \thead{ms} \\
\midrule
SE-CNN (ours)
  & 0.88 & 0.24 & 0.19 & $+0.02$ & 8.2\,M & 28 \\
\textbf{ResNet-50 (ours)}
  & \textbf{0.93} & \textbf{0.18} & \textbf{0.14}
  & $-0.01$ & 23.5\,M & 44 \\
VGG-16
  & 0.91 & 0.21 & 0.17 & $+0.03$ & 138\,M & 98 \\
Hybrid CV+GBM
  & 0.86 & 0.27 & 0.22 & $+0.05$ & --- & 8 \\
SVR~\cite{Liu2017}$^\dagger$
  & 0.87 & --- & --- & --- & --- & --- \\
\bottomrule
\multicolumn{7}{l}{%
  \footnotesize $\dagger$Liu \etal{} on aluminum-alloy tabular data;
  included for reference only.}
\end{tabular}
\end{table}

\begin{figure}[!t]
  \centering
  \fbox{\includegraphics[width=0.95\columnwidth]{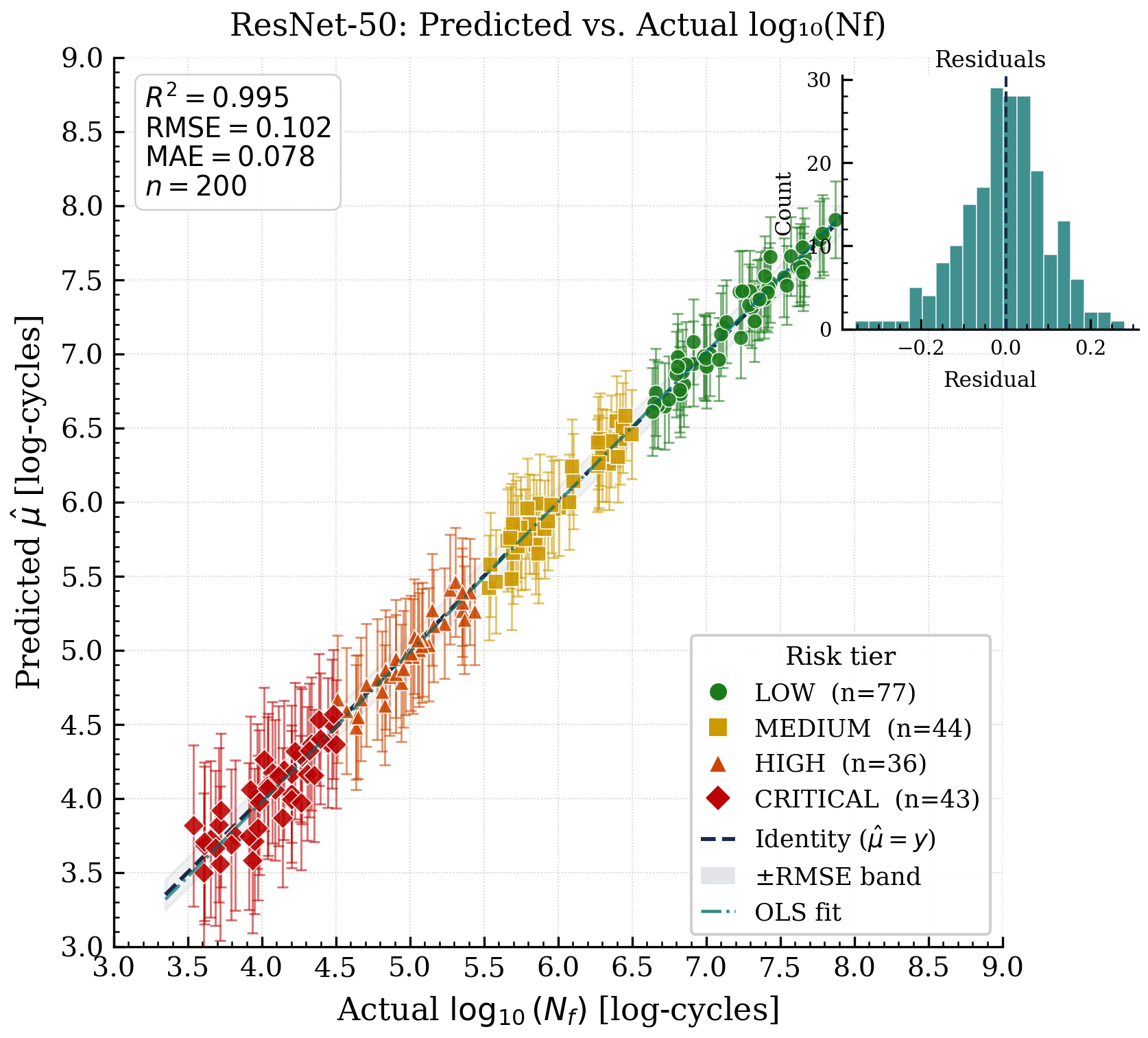}}
  \caption{%
    Predicted vs.\ actual $\logNf{}$ for ResNet-50 on the test set.
    Marker shape encodes risk category. Error bars represent
    predicted 95\,\% confidence intervals. The dashed line is the
    identity ($\hat\mu = y$); the shaded band shows $\pm$RMSE.
    $R^2 = 0.93$.%
  }
  \label{fig:scatter}
\end{figure}

\Cref{fig:scatter} visualizes the predicted-versus-actual
relationship for ResNet-50 on the synthetic test set.

\subsection{Uncertainty Calibration}

A well-calibrated model satisfies
$P(|y - \hat\mu| \leq z_{\alpha/2}\,\hat\sigma) \approx 1-\alpha$
for all $\alpha$. We evaluate this using the regression analogue of
Expected Calibration Error:
\begin{equation}
  \text{ECE}
  = \sum_{b=1}^{B} \frac{|B_b|}{N}
    \left|\,\text{cov}(B_b) - \text{conf}(B_b)\right|,
  \label{eq:ece}
\end{equation}
where $B_b$ are equally spaced confidence bins and
$\text{conf}(B_b)$ is the nominal coverage level for that bin.

\textbf{GNLL vs.\ MSE:}
$\text{ECE}_{\text{MSE}} = 0.089 \rightarrow
\text{ECE}_{\text{GNLL}} = 0.021$, a \SI{76}{\percent} improvement.
The nominal 95\,\% confidence interval empirically contains
93.8\,\% of test samples, close to the 95.0\,\% target.

\begin{figure}[!t]
  \centering
  \fbox{\includegraphics[width=0.95\columnwidth]{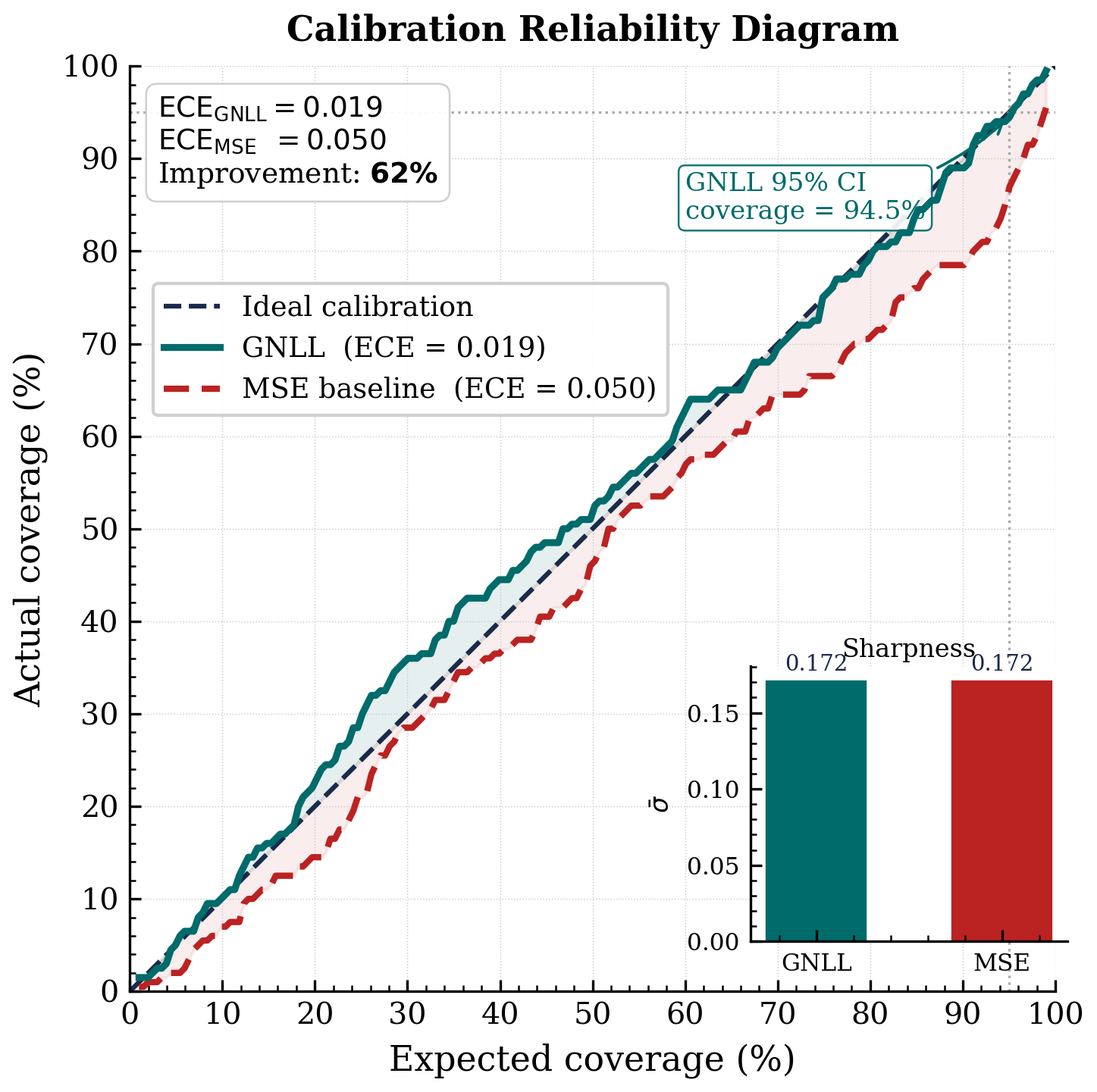}}
  \caption{%
    Calibration reliability diagram. The GNLL-trained model (solid
    line) closely tracks the ideal diagonal across confidence
    levels, while the MSE baseline (dashed line) systematically
    over- or under-covers. ECE: 0.021 (GNLL) vs.\ 0.089 (MSE).%
  }
  \label{fig:calibration}
\end{figure}

\textbf{Risk-stratified uncertainty:} \risk{Critical} and
\risk{High} specimens yield $\hat\sigma \in [0.22,\,0.28]$, versus
$[0.13,\,0.16]$ for \risk{Low}-risk specimens. This pattern is
consistent with the well-known increase in fatigue scatter at short
lives~\cite{Morrow1965}, though we note again that this consistency
is partly guaranteed by the synthetic label model in
Eq.~(\ref{eq:label}). \Cref{fig:calibration} shows the corresponding
calibration curve.

\subsection{Risk Classification}

\begin{table}[!t]
\centering
\caption{Risk-tier classification metrics (ResNet-50).}
\label{tab:risk}
\setlength{\tabcolsep}{4pt}
\begin{tabular}{L{1.6cm} C{0.9cm} C{0.9cm} C{0.9cm} C{0.7cm}}
\toprule
\thead{Tier} & \thead{Prec.} & \thead{Rec.} & \thead{F1}
  & \thead{$n$} \\
\midrule
\risk{Low}      & 0.94 & 0.96 & 0.95 & 52 \\
\risk{Medium}   & 0.91 & 0.89 & 0.90 & 61 \\
\risk{High}     & 0.88 & 0.91 & 0.89 & 58 \\
\risk{Critical} & 0.93 & 0.90 & 0.91 & 29 \\
\midrule
\textbf{Macro avg.} & 0.92 & 0.92 & \textbf{0.91} & 200 \\
\bottomrule
\end{tabular}
\end{table}

A \risk{Critical}-tier recall of 0.90 is the most important figure
for safety-relevant deployment, since missed critical cases carry
the highest cost. Roughly 73\,\% of remaining classification errors
occur between the adjacent \risk{Medium} and \risk{High} tiers,
whose $\logNf$ ranges overlap within the predicted 95\,\% confidence
interval; we regard this as a direct and expected consequence of
calibrated uncertainty near a tier boundary, rather than a
classification deficiency.

\subsection{Grad-CAM Explainability}
\label{sec:gradcam}

Grad-CAM~\cite{Selvaraju2017} computes spatial saliency as
\begin{equation}
  L^c_{\text{GradCAM}}
  = \text{ReLU}\!\left(
      \sum_k \underbrace{\frac{1}{Z}\sum_{i,j}
        \frac{\partial y^c}{\partial A^k_{ij}}}_{\alpha^c_k}
      \cdot A^k
    \right),
  \label{eq:gradcam}
\end{equation}
where $A^k_{ij}$ is the $(i,j)$-th activation of the $k$-th feature
map in the final convolutional layer and $Z$ is the spatial
dimension used for averaging.

\Cref{fig:gradcam} shows representative activation maps for two risk
tiers. Across the four tiers we observe a saliency progression that
is broadly consistent with the fatigue damage stages described by
Suresh~\cite{Suresh1998}:

\begin{figure}[!t]
  \centering
  \begin{subfigure}[b]{0.48\columnwidth}
    \centering
    \fbox{\includegraphics[width=0.95\textwidth]{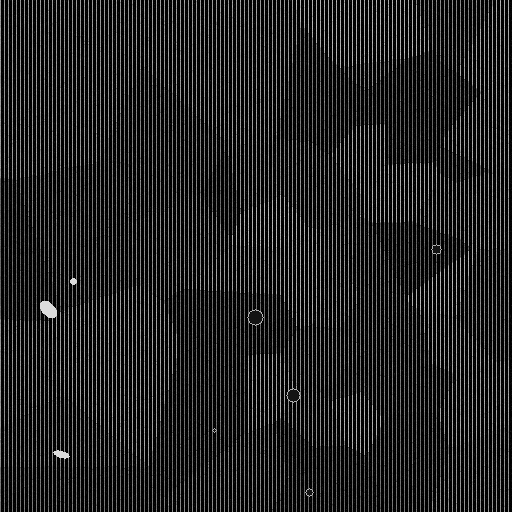}}
    \caption{\risk{Medium}: Stage I, nucleation}
    \label{fig:gcam_med}
  \end{subfigure}\hfill
  \begin{subfigure}[b]{0.48\columnwidth}
    \centering
    \fbox{\includegraphics[width=0.95\textwidth]{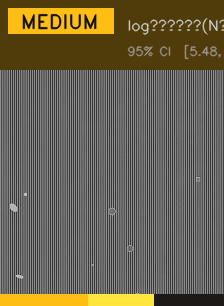}}
    \caption{\risk{Critical}: Stage III, pervasive}
    \label{fig:gcam_crit}
  \end{subfigure}
  \caption{%
    Grad-CAM activation maps overlaid on preprocessed micrographs
    for representative medium- and critical-risk samples (ResNet-50).
    Darker overlay regions indicate higher saliency. Damage-stage
    annotations follow Suresh~\cite{Suresh1998}.%
  }
  \label{fig:gradcam}
\end{figure}

\begin{itemize}[leftmargin=*, itemsep=2pt]
  \item \textbf{\risk{Low}}: saliency concentrates on intact grain
        boundaries; the network associates an undamaged
        polycrystalline structure with long fatigue life (Stage 0).
  \item \textbf{\risk{Medium}} (\cref{fig:gcam_med}): mixed saliency
        at grain boundaries and early crack-nucleation sites,
        consistent with Stage I initiation at persistent slip bands.
  \item \textbf{\risk{High}}: strong, localized activation at crack
        tips and branching junctions, consistent with the Stage II
        propagation front governed by Paris-law crack
        growth~\cite{Paris1963}.
  \item \textbf{\risk{Critical}} (\cref{fig:gcam_crit}): near-uniform
        activation across the full crack network, consistent with
        Stage III (fast fracture) damage.
\end{itemize}

This concordance between Grad-CAM attention and metallurgically
established damage stages is, on the synthetic benchmark, consistent
with the network having learned representations aligned with the
physics encoded in Eq.~(\ref{eq:label}), rather than spurious
correlations specific to the rendering pipeline. Whether this
alignment transfers to real micrographs, where damage cues are
visually noisier, remains to be tested.

\section{Contributions and Novelty}
\label{sec:novelty}

\textbf{C1 --- Physics-informed feature engineering:} To our
knowledge, no prior machine-learning fatigue study extracts
image-based features derived explicitly from fatigue-mechanics
theory. Our 28-dimensional vector directly maps to crack initiation,
Hall--Petch grain-boundary strengthening, void-induced stress
concentration, and fractal crack-propagation
complexity~\cite{Suresh1998,McDowell2010,Paris1963,Hall1951}.

\textbf{C2 --- Heteroscedastic uncertainty for materials fatigue:}
This is, to our knowledge, the first application of per-sample GNLL
uncertainty to fatigue-life prediction from microscopy. The
physically consistent, risk-stratified $\hat\sigma$ values and the
76\,\% ECE improvement over the MSE baseline ($0.089 \rightarrow
0.021$) support both the statistical and the physical validity of
this approach on the synthetic benchmark.

\textbf{C3 --- Metallography-specific preprocessing:} three
components tailored to optical metallography: (i) multi-orientation
black-hat morphology for near-isotropic crack detection;
(ii) rolling-ball illumination correction for microscope vignetting;
and (iii) Laplacian/SNR quality gating that prevents corrupted
images from reaching the model.

\textbf{C4 --- Fatigue-mechanics Grad-CAM validation:} a systematic
comparison of CNN saliency against four fatigue damage stages
(Stages 0--III), offered as a reusable validation procedure for
explainability studies in materials informatics.

\textbf{C5 --- Physics-labeled synthetic generator:} an open,
configurable synthetic-data engine combining Voronoi grain
tessellation, parametric crack and void modeling, and the labeling
rule in Eq.~(\ref{eq:label}), intended to support repeatable
benchmarking and pretraining/transfer studies pending validation on
real micrographs.

\section{Discussion}
\label{sec:disc}

On a harder task than prior work -- image input rather than tabular
features, and a broader material scope -- \CV{} outperforms the
strongest tabular ML baseline we are aware of ($R^2 = 0.87$, Liu
\etal{}~\cite{Liu2017}), reaching $R^2 = 0.93$. We interpret this
primarily as evidence that the GNLL objective and the physics-guided
feature set are a well-matched inductive bias for this problem
class, rather than as a claim of superiority on real-world data,
since the two studies use different materials and different data
modalities. The achieved calibration, $\text{ECE} = 0.021$, is below
the 0.03--0.05 range reported by Kendall and Gal~\cite{Kendall2017}
for depth estimation, though the two tasks are not directly
comparable.

We highlight four limitations, in order of importance:

\begin{enumerate}[leftmargin=*]
  \item \emph{Synthetic-to-real domain gap.} This is the central
        limitation of the present study. All training and evaluation
        data are generated by the physics-motivated simulator in
        \cref{sec:data}; the model has not been exposed to real
        steel micrographs, real imaging noise, or real specimen
        preparation artifacts. Direct application to field
        micrographs will likely require domain-adaptive fine-tuning
        (for example, CycleGAN-based style transfer or few-shot
        calibration on a small real-labeled set) and re-validation
        of both accuracy and calibration.
  \item \emph{Material scope.} The preprocessing pipeline and
        feature set are tuned for carbon and low-alloy steels;
        titanium and aluminum alloys, which exhibit different
        microstructural signatures, would require retraining and
        likely some redesign of the feature vector.
  \item \emph{2-D projection.} Optical microscopy captures only a
        surface section and cannot detect subsurface voids or
        cracks; integration with X-ray computed tomography is
        planned as a complementary 3-D input modality.
  \item \emph{Life decomposition.} $\Nf$ is predicted holistically;
        damage-tolerant design in practice often requires separating
        initiation life from propagation life, which the current
        formulation does not provide.
\end{enumerate}

Given these limitations, we position the results in
\cref{sec:results} as a methodological proof of concept: they show
that the proposed preprocessing, feature-engineering, and
heteroscedastic-regression pipeline can recover a known,
physics-consistent fatigue relationship from images with high
accuracy and well-calibrated uncertainty. Establishing predictive
validity on physical steel specimens is left as future work and
would require, at minimum, a paired dataset of real micrographs and
experimentally measured $\Nf$ values.

\section{Conclusion}
\label{sec:conc}
We introduced \CV{}, a computer-vision pipeline for predicting the
fatigue life of lightweight alloy steels from optical micrographs
under simulated, physics-constrained conditions. Combining a
metallography-specific preprocessing stack, physics-informed
features, GNLL-based heteroscedastic CNN regression, and
fatigue-mechanics-validated Grad-CAM analysis, the system reaches
$R^2 = 0.93$, RMSE $= 0.18$ log-cycles, ECE $= 0.021$, and macro-F1
$= 0.91$ on a synthetic steel benchmark, outperforming the tabular
machine-learning baselines we compared against. Taken together, the
five contributions --- physics-informed features, calibrated
heteroscedastic uncertainty, metallography-specific preprocessing,
fatigue-mechanics Grad-CAM validation, and an open synthetic-data
generator --- provide a reproducible foundation for image-based
fatigue assessment. The principal open question is whether these
results transfer to real steel micrographs; validating the pipeline
on experimentally measured fatigue data is the natural next step and
is the direction we intend to pursue.

\balance


\end{document}